\documentclass[10pt,twocolumn,letterpaper]{article}

\usepackage[algorithms]{wacv}      

\usepackage{graphicx}
\usepackage{amsmath}
\usepackage{amssymb}
\usepackage{booktabs}
\usepackage{algorithm}
\usepackage{algorithmicx} 
\usepackage{algpseudocode}
\usepackage{amsmath}

\usepackage{tabularx}
\usepackage{float}

\usepackage[pagebackref,breaklinks,colorlinks]{hyperref}

\usepackage[capitalize]{cleveref}
\crefname{section}{Sec.}{Secs.}
\Crefname{section}{Section}{Sections}
\Crefname{table}{Table}{Tables}
\crefname{table}{Tab.}{Tabs.}

\def\wacvPaperID{468} 
\def\confName{WACV}
\def\confYear{2024}

\begin{document}

\title{CPSeg: Finer-grained Image Semantic Segmentation via Chain-of-Thought Language Prompting}

\author{Lei Li\\
University of Copenhagen\\
\\
{\tt\small lilei@di.ku.dk}
}
\maketitle

\begin{abstract}

Natural scene analysis and remote sensing imagery offer immense potential for advancements in large-scale language-guided context-aware data utilization. This potential is particularly significant for enhancing performance in downstream tasks such as object detection and segmentation with designed language prompting. In light of this, we introduce the \textbf{CPSeg} (\textbf{C}hain-of-Thought Language \textbf{P}rompting for Finer-grained Semantic \textbf{Seg}mentation), an innovative framework designed to augment image segmentation performance by integrating a novel "Chain-of-Thought" process that harnesses textual information associated with images. This groundbreaking approach has been applied to a flood disaster scenario. \textbf{CPSeg} encodes prompt texts derived from various sentences to formulate a coherent chain-of-thought. We propose a new vision-language dataset, FloodPrompt, which includes images, semantic masks, and corresponding text information. This not only strengthens the semantic understanding of the scenario but also aids in the key task of semantic segmentation through an interplay of pixel and text matching maps. Our qualitative and quantitative analyses validate the effectiveness of \textbf{CPSeg}.

\end{abstract}
%
%

Image segmentation has emerged as a critical component in the analysis of remote sensing imagery, aiming to partition an image into multiple segments, or sets of pixels, that correspond to distinct objects or object components \cite{xie2021segformer,li2023edge,brandt:20, li2022mask,zhang2020fact, zhang2023attention}. Its significance is further magnified in the context of remote sensing as a global observation system, encompassing various applications such as urban planning, resource management \cite{oehmcke2022deep,revenga2022above,LR-CS-10065722}, environmental monitoring \cite{li2023segment,zhou2023multi,li2022buildseg}, and particularly, disaster response \cite{rahnemoonfar2021floodnet}. However, accurately segmenting remote sensing images is confronted with complexities arising from diverse textures, irregular shapes, and varying scales present in these images. Consequently, the development of effective segmentation methods remains an ongoing challenge and a research priority.

\begin{figure}[t]
    \centering
    \includegraphics[width=0.46\textwidth]{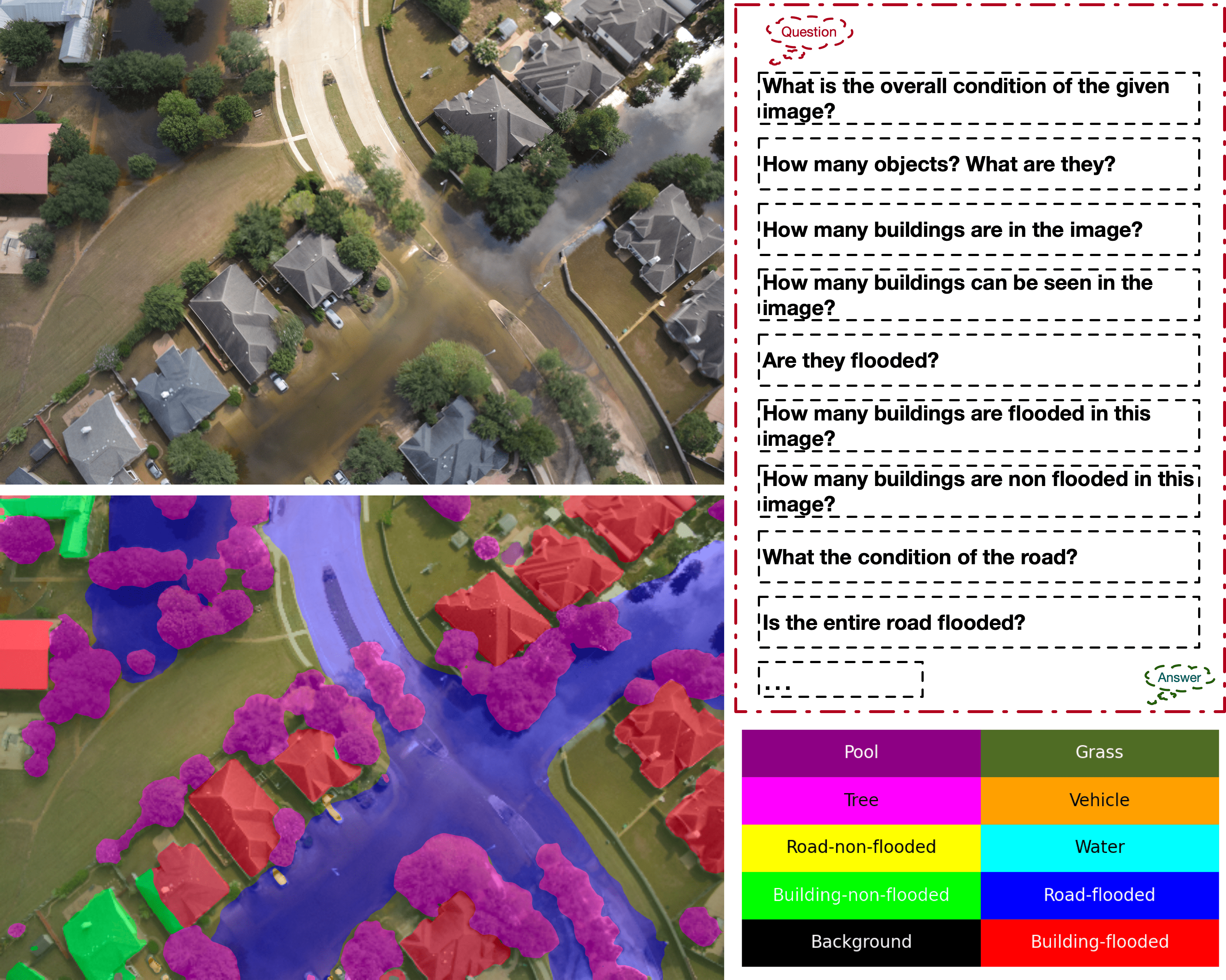}
    \caption{In the visualization of fine-grained image semantic segmentation, the incorporation of chain-of-thought language prompting proves instrumental in attaining meticulous and accurate outcomes. The left panel of the illustration showcases a bird's-eye view, juxtaposed with the output mask derived from the application of CPSeg to the original image. This arrangement offers a comprehensive outlook on the original image and its corresponding segmentation. Conversely, the right panel emphasizes the chain-of-thought prompting process, employing diverse question types. It showcases color-coded masks that represent each thought-provoking question in the chain, accompanied by their finer-grained segmentation outputs. This dual-paneled visualization approach effectively portrays the intricate procedure and intricate outcomes of image semantic segmentation through the utilization of chain-of-thought language prompts.}
    \label{fig:motivation}
\end{figure}

Semantic segmentation is a critical task in computer vision that involves dense pixel-wise classification. Drawing inspiration from the success of contrastive VL (Vision-Language) pre-training, specifically CLIP~\cite{radford2021learning}, several recent works~\cite{li2022language, yun2023ifseg, rao2022denseclip, xu2022simple} have explored CLIP-based segmentation approaches to improve the transfer of language features and enhance segmentation performance. VL segmentation methods~\cite{ghiasi2022scaling,li2022language,xu2022simple} also face challenges in terms of training with additional image data, acquiring segmentation annotations, or obtaining natural language supervision. These challenges are crucial when adapting pre-trained vision-language models to downstream segmentation tasks, and they significantly impact the optimization and scalability of these models.

Existing methodologies for VL based image semantic segmentation often overlook the incorporation of human cognitive processes and sequential thought patterns \cite{wei2022chain}, particularly within the field of remote sensing where such approaches are scarce. The work of Wei et al. \cite{wei2022chain} introduced a chain-of-thought network that demonstrated the effectiveness of this approach in natural language processing. However, there has been limited exploration of this methodology in the context of image segmentation. This gap in research becomes evident when analyzing complex images, such as flood scenes \cite{rahnemoonfar2021floodnet}, as depicted in Figure ~\ref{fig:motivation}. A human observer naturally engages in a tiered process of analysis, initially identifying distinct classes within the image (e.g., buildings or roads), followed by evaluating the quantity and extent of their impact due to flooding (e.g., number of submerged buildings or impassable roads). Unfortunately, this sequential and logical cognitive process remains largely unexplored in existing image semantic segmentation frameworks. This presents an opportunity to enhance these models by integrating insights from human cognitive processing.

Meanwhile, the progress in remote sensing technologies has opened up intriguing possibilities for enhancing image segmentation methodologies. In this study, we aim to investigate this potential by introducing a novel framework for image segmentation that harnesses language-guided context-aware data, an approach that has been widely utilized in the analysis of natural scenes but remains relatively unexplored in the domain of remote sensing imagery. Our experimental results have substantiated the effectiveness of the proposed framework, specifically in its utilization of a chain-of-thought process to iteratively incorporate textual information into the image segmentation pipeline.

Our work explore to address this gap by introducing a novel framework that utilizes a chain-of-thought continual-vision strategy to enhance image segmentation in remote sensing. While several studies have explored different strategies to improve image segmentation accuracy, our approach focuses on leveraging the chain-of-thought process, which involves sequential reasoning to analyze complex images. By mimicking human cognition, this approach enables the sequential thought process that humans employ to identify, relate, and understand various elements within an image. Our framework integrates textual information derived from images in a continuous manner, effectively leveraging language data to enhance segmentation performance. This approach is particularly beneficial in time-critical scenarios such as disaster response, where the ability to not only identify but also understand the spatial relationships and context of objects through chain-of-thought processing in an image can be crucial.

Our findings demonstrate that our proposed method significantly improves segmentation outcomes in a flood disaster scenario when using the FloodPrompt for empirical validation. The utilization of a text encoder to process prompt texts from different sentences, and the incorporation of the encoded information in the semantic segmentation task, have proven to be particularly advantageous. Our contributions can be summarized as follows:
\begin{itemize}
\item We introduce a novel methodology for finer-grained image semantic segmentation specifically designed for remote sensing imagery, harnessing the power of language prompting.
\item We propose a novel task that incorporates the concept of chain-of-thought prompting into the domain of image semantic segmentation, paving the way for more advanced segmentation algorithms.
\item Through an extensive validation process, we demonstrate that our proposed FloodPrompt dataset outperforms conventional methods in terms of label semantic segmentation and language-guided approaches, highlighting its superior efficacy.
\end{itemize}


We begin this paper with a comprehensive introduction, which includes a detailed overview of our novel framework and a review of related work in the field. We then proceed to present a thorough explanation of our methodology, outlining the various components and their roles. Following this, we provide an extensive analysis of our experimental results, encompassing both quantitative and qualitative assessments. To further elucidate the contributions of different components, we also include an ablation study. Finally, we conclude the paper with a discussion on the implications of our findings and outline potential avenues for future research.

\begin{figure*}[t]
    \centering
    \includegraphics[width=0.9\textwidth]{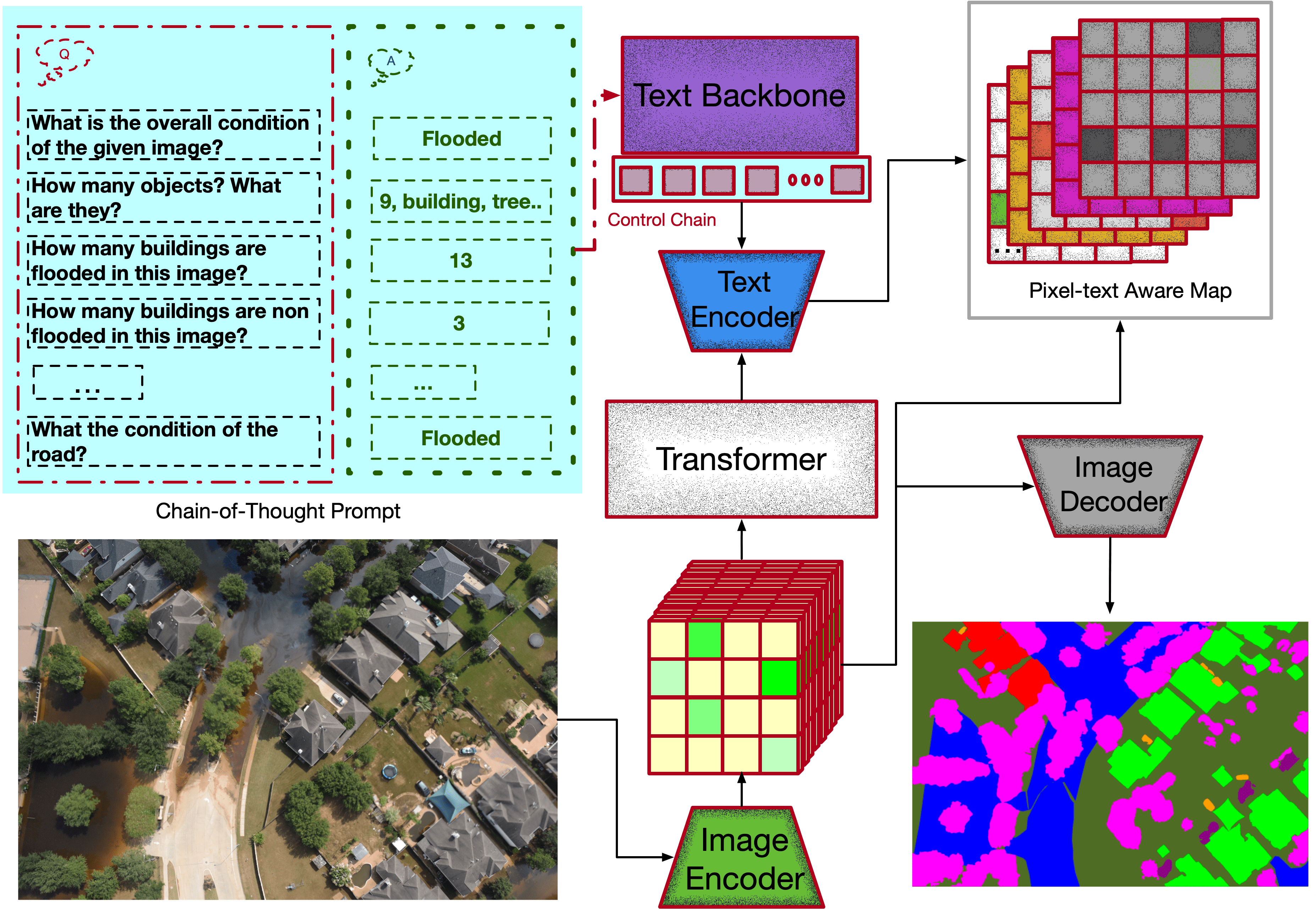}
    \caption{The CPSeg framework is designed for image semantic segmentation and involves several key steps. Firstly, it extracts embeddings from both images and chain-of-thought text prompts. These embeddings are then used to compute pixel-text aware maps, which represent a novel adaptation of CLIP's image-text matching problem. CPSeg introduces a context-aware approach that enables dense prediction in segmentation tasks. Additionally, it incorporates a transformer module \cite{vaswani2017attention} for both the text and vision backbone, leveraging pre-trained knowledge to improve the accuracy of segmentation results. This innovative methodology expands the possibilities for achieving more precise and nuanced image semantic segmentation.}
    \label{fig:Pipeline}
\end{figure*}

\section{Related Work}

\paragraph{Vision-language pre-training.} Recent advancements in vision-language models, which are pre-trained on large-scale image-text datasets, have demonstrated remarkable efficacy in adapting to novel tasks within the context of zero-shot and few-shot learning, spanning diverse domains \cite{alayrac2022flamingo,antol2015vqa,xu2015show}. These developments highlight the significant potential for applying such models in complex computational domains. Notably, models incorporating dual encoders, multi-modal encoders, and encoder-decoders, such as CLIP, have further improved cross-modal representation quality through contrastive pre-training. Concurrently, the "pre-training + fine-tuning" paradigm has revolutionized computer vision and natural language processing by initially pre-training models on extensive datasets like ImageNet \cite{deng2009imagenet}, JFT \cite{sun2017revisiting}, and Kinetics \cite{carreira2017quo}, followed by fine-tuning for various downstream tasks. This framework often evolves into a prompt-based paradigm, wherein downstream tasks are reformulated to align with those addressed during the pre-training process. Collectively, these strategies reflect the evolving landscape of vision-language pre-training, holding promise for advancing performance in complex tasks.

\paragraph{Image segmentation with Vision-language.}
Image segmentation~\cite{long2015fully,zhao2017pyramid,lin2017feature,chen2017deeplab} remains a central yet challenging task in computer vision, particularly when segmenting novel visual categories. A variety of approaches have been explored, including unsupervised, zero-shot segmentation, and methods leveraging vision-language models. Unsupervised segmentation approaches often focus on clustering dense image representations and matching these to corresponding segmentation categories, while vision-language (VL)~\cite{liu2022open,zhou2021denseclip} driven strategies aim to replace the matching process with text encoders for enhanced efficiency and transferability. Meanwhile, transferring methods~\cite{bucher2019zero} often necessitate class-agnostic or class-specific segmentation annotations, despite recent innovations using VL models. In the context of these developments, this paper explores image-free semantic segmentation, aiming for practical applicability in scenarios where only segmentation vocabulary is given, providing a simpler alternative to collecting images or other annotations. 
To address the significant annotation burden associated with previous supervised pre-training settings, several self-supervised pre-training approaches have been introduced in the field of dense prediction~\cite{caron2021emerging,wang2021dense,rao2022denseclip} with fine-tuning strategy that harnesses the knowledge embedded within large-scale vision-language pre-trained models. Importantly, this strategy incorporates language information as a guiding component within the learning process, marking a distinct departure from traditional methodologies. The evolving intersection of computer vision and natural language processing fields, especially with vision-language pre-training, offers new perspectives for these challenges, with models like CLIP demonstrating impressive transferability over diverse classification datasets. Yet, very few attempts have been made to apply such models to image segmentation prediction tasks, making it a compelling area for future exploration.


\section{Methodology}

The CPSeg begins with the extraction of embeddings from both the input image and the chain-of-thought text prompts. These embeddings are then utilized to generate pixel-text aware maps. we adapt the chain-of-thought process to suit the specific requirements of our task, namely, improving the interpretative capability of our model with the help of corresponding textual information.

\subsection{Overview}
The CPSeg framework incorporates a dynamic calculation of the pixel-text match loss within the pixel-text aware map, which is updated by the transformer's parameters as more prompts are provided. The resulting score maps are then fed into a decoder that utilizes ground-truth labels for supervision. Moreover, CPSeg capitalizes on the wealth of pre-trained knowledge(CLIP) by leveraging contextual information present in images to guide the language model prompts. This is achieved through the integration of a transformer module, enabling the framework to optimize its understanding of the image's context and thereby improving the quality of the segmentation results. The overall architecture of CPSeg establishes it as a robust and efficient tool for image semantic segmentation.

\begin{figure*}[h]
    \centering
    \includegraphics[width=0.9\textwidth]{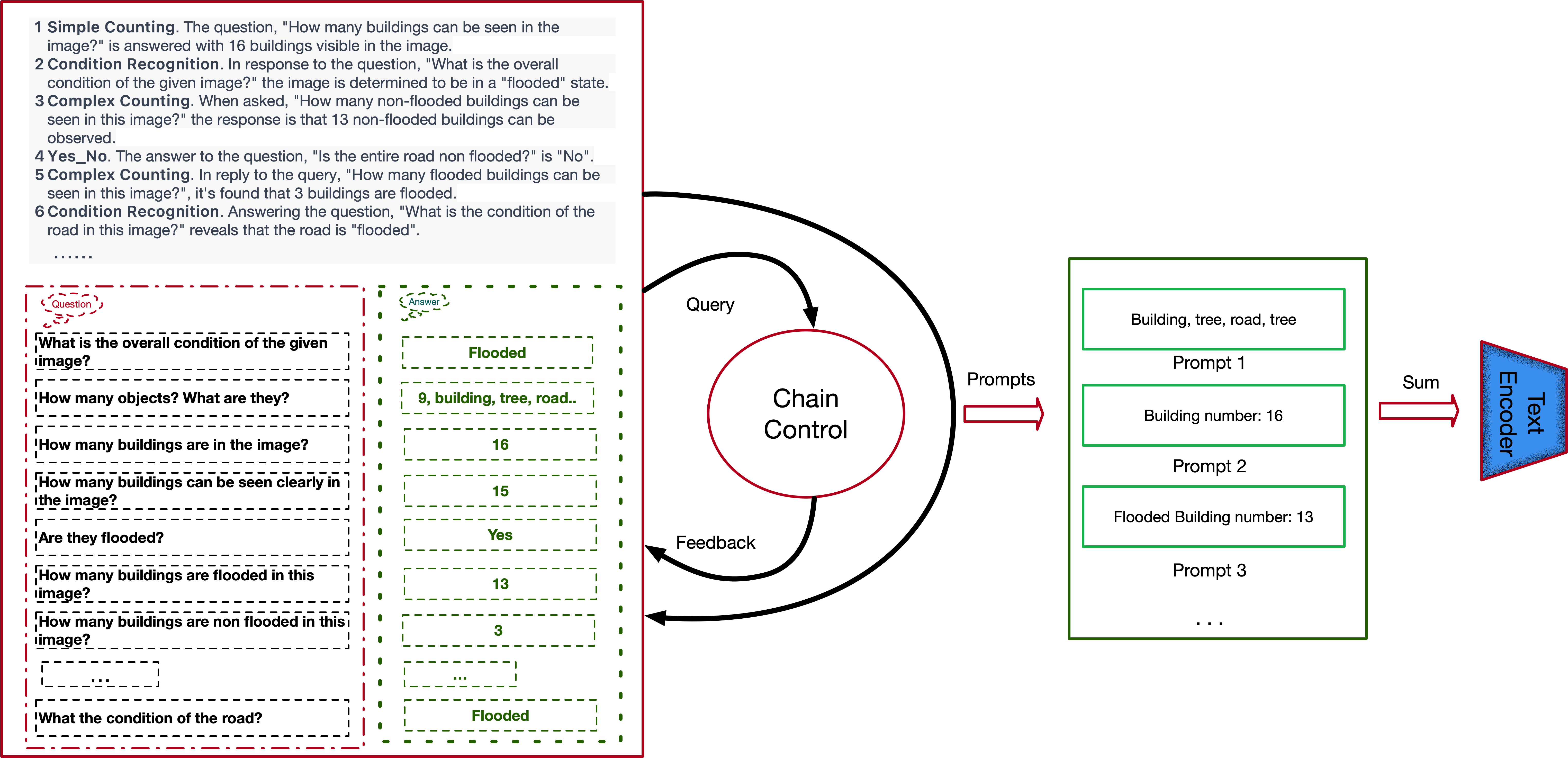}
    \caption{Our chain-of-thought prompting pipeline is intricately designed to elicit comprehensive responses. When relevant classes are identified, the pipeline proceeds to inquire about the precise numbers involved. Furthermore, it thoroughly examines all the potential questions within the context, ensuring a robust and detailed understanding.}
    \label{fig:Chain_control}
\end{figure*}

The chain-of-thought process as show in Figure~\ref{fig:Chain_control}, a crucial element of this framework, is rooted in the theoretical understanding of human cognition.
The methodology of CPSeg employs a pre-trained Vision Transformer, denoted as \( V \). It mandates various prerequisites including the total number of tasks, \( T \); a comprehensive training set, denoted as \(\{(I^t_i, L^t_i)\}_{i=1, t=1}^{N_t, T}\); a collection of prompts, \( Q \), and their respective keys, \( K \). Subsequent steps involve the determination of specifics like the number of training epochs for the \( t \)-th task, \( E_t \), the learning rate \( \eta \), and a balancing parameter \( \lambda \). The objective is to update and optimize the parameters of \( V \), \( Q \), and \( K \), using a hierarchical prompting mechanism. For more details about prompting mechanism, refer to supplement.

It breaks down complex tasks into a sequence of smaller, more manageable decisions. This offers a practical approach for handling intricate data analysis tasks such as image segmentation. Formally, given an image $I$ with pixel data $P = {p_1, p_2, \ldots, p_n}$, the chain-of-thought process handles the segmentation task as a sequence of decisions concerning each pixel $p_i$. CPSeg enhances this process by integrating context-aware data guided by language. In particular, we construct a chain of thoughts $C = {c_1, c_2, \ldots, c_m}$, where each thought $c_i$ corresponds to a sentence $s_i$ in the text accompanying the image. Each thought $c_i$ consists of a text encoder $T(c_i)$ and a pixel-level segmentation function $f_{c_i}(p_i)$, generating a sequence of segmentation decisions $D = {d_1, d_2, \ldots, d_m}$, with each decision $d_i$ corresponding to a thought $c_i$.

To support this process, we employ a text encoder $E$ that generates encoded representations from diverse sentence prompts. Formally, for a given sentence $s_i$, we derive its encoded representation $e_i = E(s_i)$, where $e_i$ captures the semantic details in $s_i$. This encoding procedure enables our framework to leverage the semantic context provided by the language data, enhancing the capability of the chain-of-thought process. The encoded data subsequently facilitates the downstream task of semantic segmentation. Each segmentation function $f_{c_i}(p_i)$ in the thought $c_i$ employs the corresponding encoded representation $e_i$ to make better-informed decisions about pixel classification. Specifically, for a given pixel $p_i$, the segmentation decision $d_i$ is computed as $d_i = f_{c_i}(p_i, e_i)$. This allows our framework to make use of both the spatial information from the pixels and the semantic information from the text, achieving more accurate and context-aware segmentation outcomes through chain-of-thought prompting. The Figure~\ref{fig:Chain_control} illustrates the flow of the chain-of-thought prompting process in the CPSeg framework.

\subsection{Language Prompting.}




The proposed framework begins with an initial pre-trained segmentation network $S$. The system is designed to handle $T$ tasks, each consisting of its own set of epochs denoted as $E_t$. At each epoch, a mini-batch is randomly sampled from the dataset, and suitable prompts and keys are generated for each image within the mini-batch. The algorithm incorporates a control mechanism that verifies the relevance of the query questions and corresponding answers. For instance, if the question relates to the number of flooded buildings, the controller specifically seeks out this information and retrieves the appropriate answer.

The framework supports various types of prompts, including simple counting and condition recognition, as depicted Figure~\ref{fig:Chain_control}. These prompts and keys are applied to the segmentation network, and the loss is calculated. Subsequently, the segmentation network is refined using gradient descent. This process is repeated for all images, mini-batches, and tasks, resulting in an updated segmentation network.

The construction of the entire chain-of-thought is hierarchical in nature, starting with macro-level considerations before delving into the specifics. In curating the prompts, we've orchestrated the keys in a hierarchical manner, commencing from overarching descriptors of the entire image and subsequently delving into more granular details, such as the presence of infrastructure elements like buildings and roads. This structured approach ensures a systematic progression from a macroscopic view to refined prompting. In the event of flood-related incidents, subsequent prompts elucidate on the inundation of infrastructure elements, quantifying inundated buildings, and further delineating the intricacies of the flooding scenario and its concomitant implications. 

For instance, the initial query might be regarding the presence of flooding, which then transitions into contemplation about the presence of buildings, and subsequently, the number of buildings inundated.


\subsection{Vision Backbone.}
The CPSeg framework tackles the challenge of semantic segmentation by adopting a Vision-and-Language (VL) encoder-decoder model. The objective is to decode a category word for each densely populated image region, considering M semantic categories of interest. However, a key challenge arises when specific semantic category words are tokenized into multiple subwords within the dictionary, introducing complexity to the task.

To overcome this challenge, we employ a Vision Transformer (ViT) as the encoder, enabling the extraction of highly detailed and efficient visual content representations. During the inference process, the encoder and decoder interact synergistically to generate the semantic segmentation mask. The decoder plays a vital role in converting the dense feature representation obtained from the encoder into category predictions for each image region. It handles the complexities arising from semantic relationships and potential subword tokenization. This approach strikes a balance between theoretical complexity and practical efficiency, providing an effective methodology for semantic segmentation tasks with reduced complications.

\begin{figure*}[ht]
    \centering
    \includegraphics[width=\textwidth]{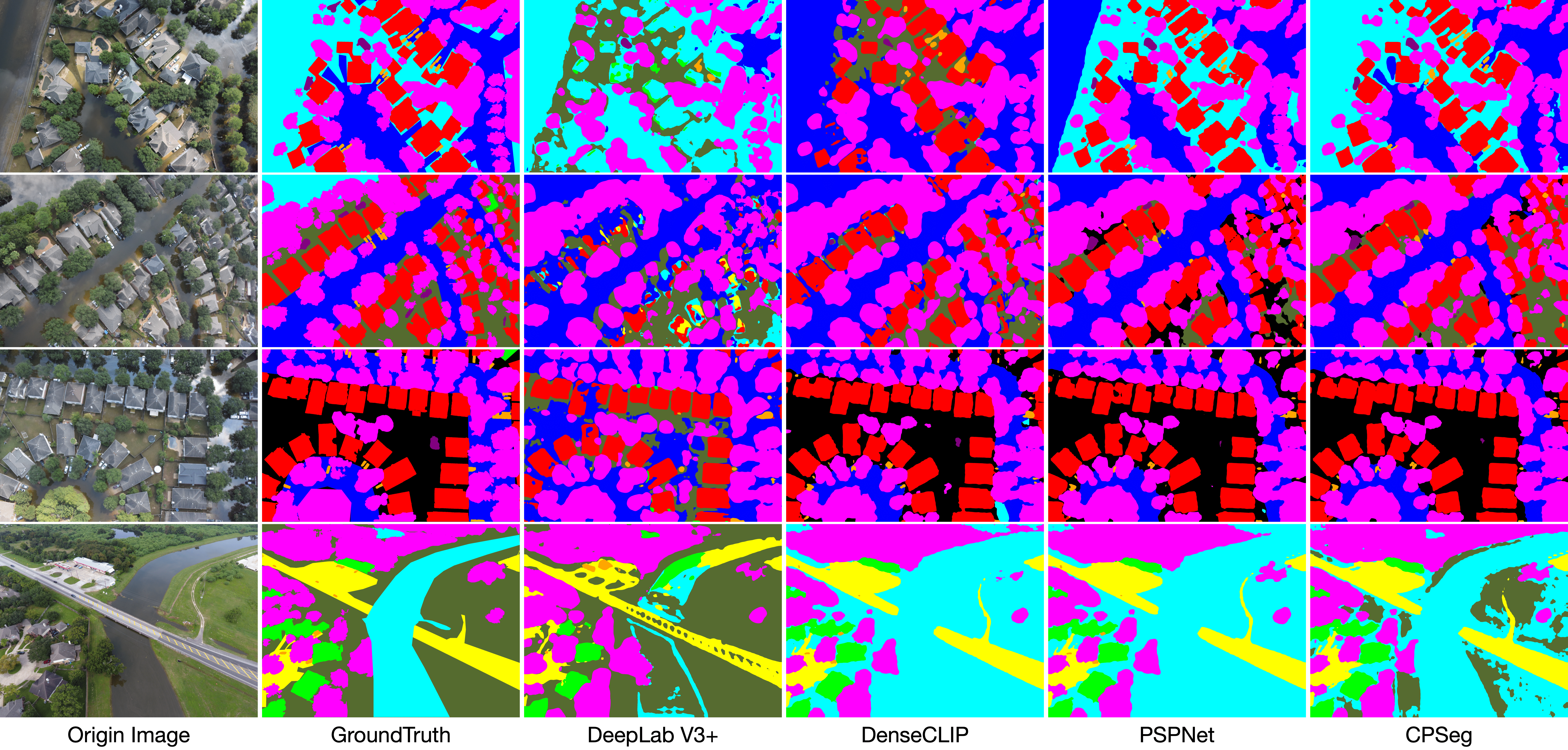}
    \caption{Visual Comparison of Segmentation Results: A comparative study of segmentation performance across DeepLab V3+, DenseCLIP, PSPNet, and our proposed CPSeg.}
    \label{fig:VisualComparison}
\end{figure*}

\subsection{Loss fuction.}

\paragraph{Pixel-Text Matching Loss.} The alignment between the image pixels and textual prompts can be quantified using the Pixel-Text Matching Loss function, mathematically denoted as $L_{\text{PTM}}$. Let us denote the set of image pixels as $P = {p_i}{i=1}^{N}$ and the set of textual prompts as $T = {t_j}{j=1}^{M}$. A similarity score $s(p_i, t_j)$, computed using an appropriate metric such as the cosine similarity or dot product, is assigned to each pair of pixel $p_i$ and text prompt $t_j$. The Pixel-Text Matching Loss is then computed as the negation of the accumulated similarity scores:

\begin{equation}
L_{\text{PTM}} = -\sum_{i=1}^{N} \sum_{j=1}^{M} s(p_i, t_j).
\end{equation}

This loss function is minimized during the model training phase, with the aim of maximizing the overall similarity between the pixels and prompts. By doing so, the model is incentivized to learn representations that improve the correspondence between similar pixels and prompts.

\paragraph{Semantic Segmentation Loss.} Our method compute score maps in segmentation. These score maps $\mathbf{s} \in \mathbb{R}^{H_4 W_4 \times K}$ can be treated as smaller-scale segmentation results. We compute a segmentation loss on them:

\begin{equation}
\mathcal{L}_{\text {seg }}=\operatorname{CrossEntropy}(\operatorname{Softmax}(\mathbf{s} / \tau), \mathbf{y}),
\end{equation}

where $\tau=0.07$ is a temperature coefficient following prior work~\cite{zhou2021denseclip}, and $\mathbf{y} \in{1, \ldots, K}^{H_4 W_4}$ denotes the ground truth labels.

\section{Experiments}
\subsection{Dataset}
To validate the efficacy of our proposed methodology, we adapted FloodNet dataset~\cite{rahnemoonfar2021floodnet} for FloodPrompt and utilized FloodPrompt for our experiments. FloodPrompt is a diverse and challenging dataset, containing a variety of remote sensing images pertinent to flood scenarios. Given the paucity of studies addressing image segmentation in such scenarios, we propose FloodPrompt provides a relevant and complex testing ground for our framework. The dataset encompasses numerous instances of flooding events captured through remote sensing technology, all annotated with detailed text descriptions, making it a suitable candidate for evaluating our language-guided segmentation approach. 

The textual descriptions associated with each image were preprocessed, tokenized, and encoded using the text encoder component of our framework. To ensure a fair comparison, the proposed method was compared with state-of-the-art segmentation methods, under identical conditions. The evaluation metrics employed for comparison included Intersection over Union (IoU), pixel accuracy, and mean accuracy, amongst others.

\begin{table*}
\centering
\begin{tabularx}{\textwidth}{l*{10}{X}}
\hline 
Method & Building Flooded & Building Non-Flooded & Road Flooded & Road Non-Flooded & Water & Tree & Vehicle & Pool & Grass & mIoU \\
\hline \hline 
ENet~\cite{paszke2016enet} & 6.94 & 47.35 & 12.49 & 48.43 & 48.95 & 68.36 & 32.26 & 42.49 & 76.23 & 42.61 \\
DeepLabV3+~\cite{chen2018encoder} & 32.7 & 72.8 & 52.00 & 70.2 & 75.2 & 77.00 & 42.5 & 47.1 & 84.3 & 61.53 \\
SegFormer-B0 ~\cite{xie2021segformer} & 70.81 & 79.04 & 69.09 & 85.27 & 80.86 & 86.06 & 56.02 & 66.13 & 91.06 & 76.20 \\
PSPNet ~\cite{zhao2018psanet} & 68.93 & 89.75 & 82.16 & 91.18 & 92.00 & 89.55 & 46.15 & 64.19 & 93.29 & 79.69 \\
DenseCLIP \cite{rao2022denseclip} & 72.98 & 79.55 & 65.94 & 84.48 & 78.97 & 85.74 & 55.01 & 63.74 & 90.92 & 75.14 \\
CPSeg & \textbf{75.54} & \textbf{92.12} & \textbf{83.46} & \textbf{91.24} & \textbf{92.01} & \textbf{93.21} & 48.01 & 64.15 & \textbf{94.21} & \textbf{82.43} \\
\hline
\end{tabularx}

\caption{Per-class results with IOU and mIoU on FloodPrompt testing set.}
\label{tab:qualitativeComparsion}
\end{table*}

\subsection{Results}
We base the mmsegmentation ~\cite{mmseg2020} to implement CPSeg. The results of our experiment were encouraging and provided substantial evidence in favor of our proposed method. From a quantitative perspective, our method consistently outperformed the state-of-the-art segmentation methods on all the evaluation metrics. For instance, the average IoU score for our method was significantly higher than that of other methods as shown in Figure~\ref{fig:VisualComparison}, and the segmented images showed that our method was able to accurately segment the various regions in the flood images, such as water bodies, vegetation, and urban areas. 

Beyond the intricate flood scenarios, we extended our experiments to utilize data from LoveDA~\cite{wang2021loveda}, further affirming that the image chain-of-thought approach can notably enhance image segmentation performance. Detailed data pertaining to LoveDA can be found in the supplementary materials.

A detailed qualitative analysis, as presented in Table~\ref{tab:qualitativeComparsion}, further attests to the efficacy of our proposed methodology. DeepLab V3+ and PSPNet primarily use image-based segmentation, whereas DenseCLIP combines standard CLIP prompts with image segmentation masks. Our observations underscore the value of incorporating textual descriptions into segmentation, leading to richer context-aware outcomes. For example, when a text description highlighted a flooded street, our method adeptly identified and segmented this specific region in the image, a task at which traditional segmentation methods often faltered. This success highlights the utility and impact of the chain-of-thought process and language-guided context in enhancing image segmentation performance.

\begin{figure}[t]
    \centering
    \includegraphics[width=0.48\textwidth]{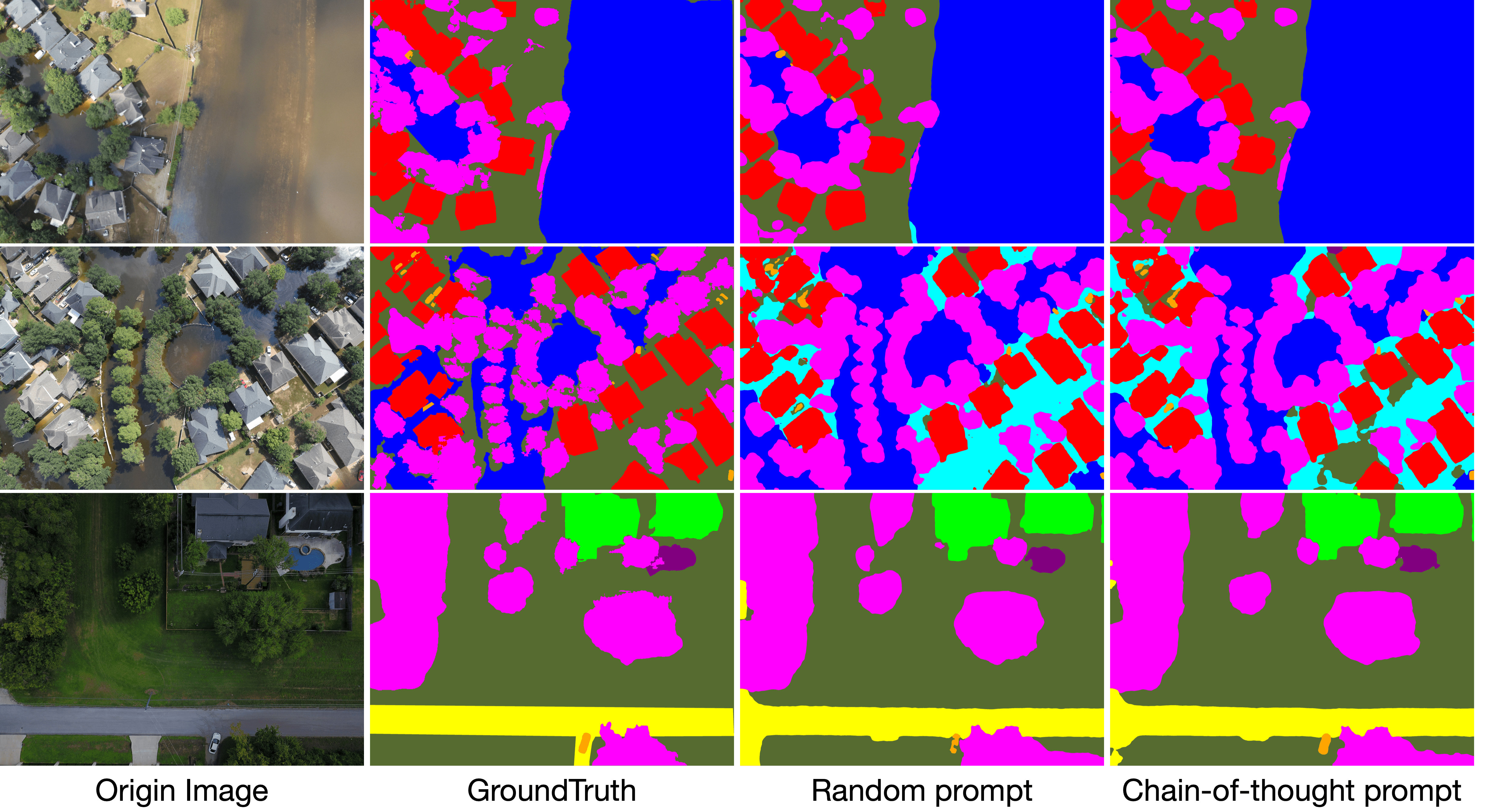}
    \caption{The performance with random prompting vs chain-of-thought prompting.}
    \label{fig:alb_}
\end{figure}

In conclusion, our experimental results unequivocally demonstrated the superiority of our proposed chain-of-thought, language-guided context-aware segmentation method over traditional image segmentation methods. By leveraging the power of language data, we were able to enhance the performance of image segmentation, particularly in the challenging and relatively unexplored domain of remote sensing imagery in flood scenarios.

\section{Ablation Study}
Firstly, we evaluated the impact of the chain-of-thought process. To do this, we compared the performance of our full model with a version that excluded the chain-of-thought process. We experiment different prompts, which standard prompt,  random prompt, two prompts, and chain-of-thought prompt, with the mIoU. The objective was to quantify the effect of sequentially injecting textual information into the image segmentation process. Our results in Table~\ref{tab:alb_prompts} showed a substantial decrease in segmentation performance when the chain-of-thought process was omitted. The associated standard and random prompts provide only a macro-level cue, with the latter randomly selecting one of the available prompts.

Specifically, the IoU score is higher indicating that the chain-of-thought process indeed plays a vital role in improving segmentation precision. This affirmed our hypothesis that a continual information stream could enhance the understanding of the scene, leading to improved segmentation outcomes.

\begin{table}[]
    \centering
        \begin{tabular}{l|c}
        \hline Prompts & mIoU $\uparrow$  \\
        \hline Standard prompt &75.26    \\
        Two prompts & 75.98   \\
        Random prompt & 76.23   \\
        Chain-of-thought prompt & 82.43  \\
        \hline
        \end{tabular}
    \caption{Ablation study with various prompting learning.}
    \label{tab:alb_prompts}
\end{table}

\begin{table*}[h]
    \centering
        \begin{tabular}{lccc}
        \hline 
        Object Class & Images (Flooded/Non-Flooded) & Images (Total) & Instances (Total) \\
        \hline \hline 
        Building & 275/1272 & (275+1272) & (3573+5373) \\
        Road & 335/1725 & (335+1725) & (649+3135) \\
        Vehicle & -/1105 & 1105 & 6058 \\
        Pool & -/676 & 676 & 1421 \\
        Tree & -/2507 & 2507 & 25889 \\
        Water & -/1262 & 1262 & 1784 \\
        \hline
        \end{tabular}
    \caption{For experimental purposes in our CPSeg pipeline, we have compared the number of finer-grained segmentation classes and consolidated number within the same class.}
    \label{tab:alb_combine}
\end{table*}



Our FloodPrompt data is finer-grained classes segmentation tasks, we also set experiments for our network for the finer-grained semantic segmentation. We combine the non-flooded building and flooded building for a building class and non-flooded road and flooded road for the road classes. Table~\ref{tab:combine} indicate if we combine the flooded and non-flooded building and road for the same classes, building, and road. Our CPSeg still work for the combined labels with the chain-of-thought prompting with implicit learning.

\begin{table}[]
    \centering
        \begin{tabular}{l|c}
        \hline Data Type & mIoU $\uparrow$  \\
        \hline 
        Original Data & 82.43 \\
        Combined Data &  87.89 \\
        \hline
        \end{tabular}
    \caption{Segmentation analysis with different type data.}
    \label{tab:combine}
\end{table}

\begin{table}[]
    \centering
        \begin{tabular}{l|c|c|c}
        \hline Method & FLOPs(G) & Params(G) & Inf time (fps)\\
        \hline 
        DenseCLIP & 1043.1 & 105.3 & 44.56\\
        CPSeg &  1037.4 & 100.8 & 42.85\\
        \hline
        \end{tabular}
    \caption{Performance with baseline for segmentation analysis.}
    \label{tab:latency}
\end{table}

In addition to performance comparisons, we analyzed the computational efficiency of our method relative to the baseline DenseCLIP, which also employs a Vision-Language model. Experiments were conducted on an NVIDIA A100 GPU, processing images of size \(1024 \times 1024\). Our model, CPSeg, exhibits lower computational complexity in terms of both parameters and floating point operations per second (FLOPs), as shown in the Table~\ref{tab:latency}. The recorded inference time for CPSeg is 42.85 FPS. These findings underscore the advantage of our chain-of-thought approach in handling finer-grained semantic segmentation tasks, offering not only superior performance but also increased efficiency compared to existing methods.

In conclusion, our ablation study provided valuable insights into the functioning of our proposed method. The results clearly showed that both the chain-of-thought process and the text encoder are crucial for our method's superior performance, thus justifying their inclusion in the framework. Furthermore, our study served to emphasize the significance of a detailed component-wise analysis in understanding and refining complex methodologies.

\section{Discussion}
While our research yields promising results, it is critical to recognize its constraints. Our experiments are predicated on a specialized dataset, targeting a specific disaster scenario - floods. Therefore, the efficacy of the proposed CPSeg method might differ with varying dataset and disaster contexts, warranting exploration in diverse scenarios like forest fires or earthquakes. Additionally, our current chain-of-thought process predominantly relies on textual cues. Hence, integrating other forms of context-sensitive data, including spatial or temporal aspects, might enhance the model's performance.

Strategically mapping these components, from an overarching view down to nuanced intricacies, is of paramount importance in this domain. As we progress, the pursuit of integrating multi-modal facets via the 'chain of thought' framework holds substantial promise. Refinement of the chain-of-thought process could potentially yield significant improvements in segmentation performance. Further, the incorporation of diverse forms of contextual data could augment the versatility and robustness of our framework. There also lies the intriguing possibility of exploring zero-shot learning in conjunction with chain-of-thought prompts from pre-trained Vision-and-Language models. Contrary to the conventional CLIP-based methodologies for image segmentation, our approach underscores the profound learning of image representation, which is achieved through the systematic provision of progressively granular prompts. Lastly, investigating the application of CPSeg other various scenarios could not only validate its effectiveness across contexts but also highlight areas requiring further improvement.

\section{Conclusion}

In conclusion, this study introduces CPSeg, a pioneering approach that employs a novel "Chain-of-Thought" process, leveraging language prompting to achieve finer-grained semantic segmentation in the field of remote sensing imagery. This innovative methodology, with particular applicability in flood disaster scenarios, capitalizes on the textual descriptions associated with images to enhance semantic understanding and improve segmentation performance. Furthermore, a new vision-language dataset, named FloodPrompt, is introduced to facilitate the evaluation of CPSeg. Through comprehensive validation, CPSeg demonstrates remarkable efficacy, pushing the boundaries of large-scale, context-aware data utilization and opening up new avenues for advancements in this domain. The results of this study offer valuable insights and contribute to the growing body of research on vision-language integration and its applications in remote sensing.
The generalized methodology is amenable to application in alternative image scenarios, provided that comprehensive contextual details are available. We believe the approach based on chain-of-thought ideas can greatly enhance high-level tasks such as segmentation, detection, and regression.
\newpage

{\small
\bibliographystyle{ieee_fullname}
\bibliography{egbib}
}

\end{document}